\documentclass{article}
\usepackage{spconf,amsmath,graphicx}

\title{Self-Learning to Detect and Segment Cysts in Lung CT Images without Manual Annotation}
\name{Ling~Zhang$^{1}$, Vissagan~Gopalakrishnan$^{2}$, Le~Lu$^{1}$, Ronald~M.~Summers$^{1}$, Joel~Moss$^{2}$, Jianhua~Yao$^{1}$
\thanks{This research was supported - in part - by the Intramural Research Program of the National Institutes of Health Clinical Center. The authors thank Dr. Li Zhang from Beijing Institute of Big Data Research for his inspiring discussion, and Nvidia for the TITAN X Pascal GPU donation. }}
\address{$^{1}$Radiology and Imaging Sciences Department,
$^{2}$Cardiovascular and Pulmonary Branch, NHLBI, \\
National Institutes of Health (NIH), Bethesda MD}

\begin{document}
%
\maketitle
\begin{abstract}
Image segmentation is a fundamental problem in medical image analysis. In recent years, deep neural networks achieve impressive performances on many medical image segmentation tasks by supervised learning on large manually annotated data. However, expert annotations on big medical datasets are tedious, expensive or sometimes unavailable. Weakly supervised learning could reduce the effort for annotation but still required certain amounts of expertise. Recently, deep learning shows a potential to produce more accurate predictions than the original erroneous labels. Inspired by this, we introduce a very weakly supervised learning method, for cystic lesion detection and segmentation in lung CT images, without any manual annotation. Our method works in a self-learning manner, where segmentation generated in previous steps (first by unsupervised segmentation then by neural networks) is used as ground truth for the next level of network learning. Experiments on a cystic lung lesion dataset show that the deep learning could perform better than the initial unsupervised annotation, and progressively improve itself after self-learning.
\end{abstract}

\section{Introduction}
\label{sec:intro}

\begin{figure}[!t]
   \begin{center}
   \begin{tabular}{c}
   \includegraphics[width=8cm]{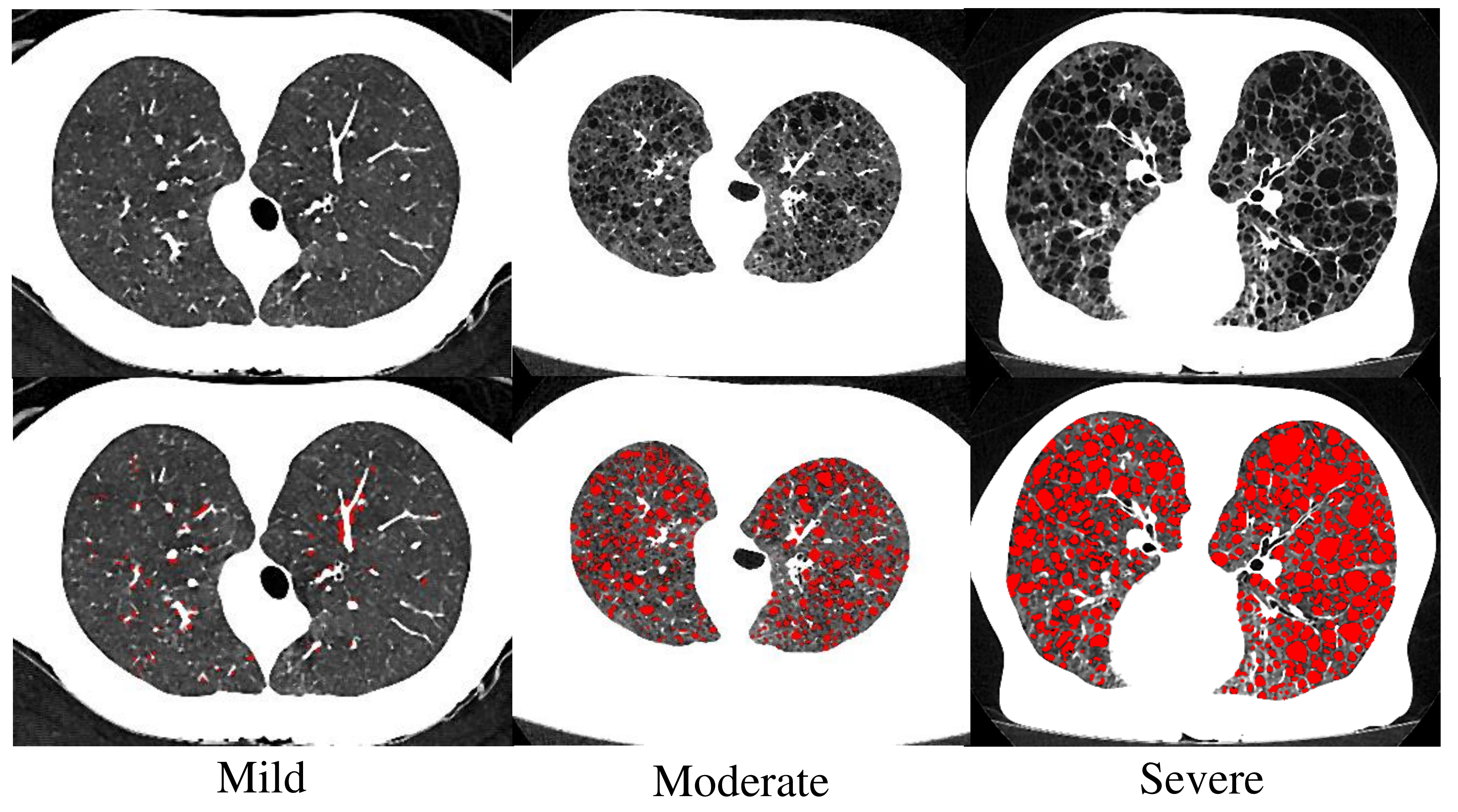}
   \end{tabular}
   \end{center}
   \caption[example]
   { \label{figannotation} 
Examples of the cystic lung lesions with different severity levels in CT image and their manual annotation (red). 
}
   \end{figure}

Image segmentation is a fundamental problem in medical image analysis. Classic segmentation algorithms \cite{sonka2014image} are usually formulated as optimization problems relying on cues from low-level image features. 
In recent years, deep learning has made much progress on image segmentation tasks (e.g., FCN \cite{long2015fully}, HED \cite{xie2015holistically}), achieved dominant performances on many medical image segmentation benchmarks, e.g., UNet \cite{ronneberger2015u} is competitive enough for many applications. 
The success of deep learning based segmentation requires supervised learning on large manually annotated data. However, expert annotations on big medical datasets are expensive to obtain or even unavailable. 
For example, 
manual annotation of hundreds of cysts in CT volume dataset (examples shown in Fig. \ref{figannotation}) is not feasible for a recent large-sized clinical study of lymphangioleiomyomatosis (LAM) \cite{yao2014sustained}. 

   \begin{figure*}[!t]
   \begin{center}
   \begin{tabular}{c}
   \includegraphics[width=14.5cm]{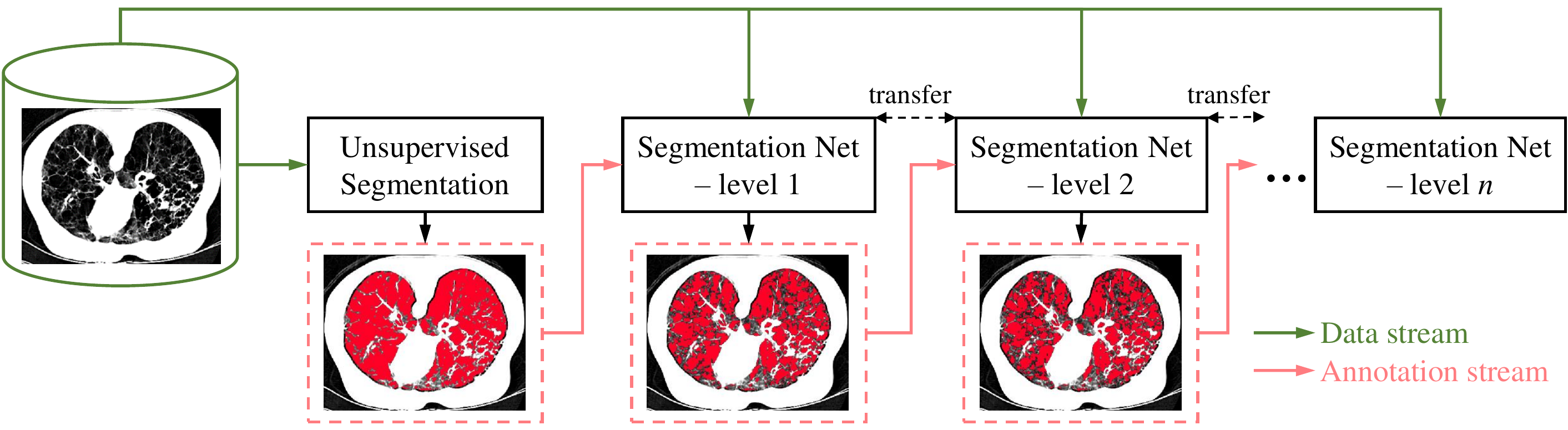}
   \end{tabular}
   \end{center}
   \caption[example]
   { \label{figframework} 
Learning to segment medical images without manual annotation. Segmentation networks (level 1 -- level $n$) are recursively trained with the previous network segmentation as training labels.
}
   \end{figure*} 

To alleviate the annotation burden, researchers exploit weakly supervised methods for deep learning based segmentation. One direction is to reduce the effort (e.g., time, expertise) for annotation. By combining FCN and active learning, 50\% training data is needed to train a model with comparable performance as training on all data \cite{yang2017suggestive}. 
Another direction applies image-level annotation by incorporating FCN in a multiple instance learning framework \cite{jia2017constrained}. However, expertise from physicians are still needed, such as assigning image-level annotations and estimating the lesion size.  

Recently, deep learning has shown a potential to beat the teacher (i.e., perform better than the training data labels) \cite{guan2017said,khoreva2017simple} or even self-learn to be an expert without human knowledge in AlphaGo Zero \cite{silver2017mastering}. Specifically, for some classification \cite{guan2017said} and semantic segmentation \cite{khoreva2017simple} tasks, when provided with data labels with certain amount of errors, deep learning could produce lower errors than the original erroneous labels. In addition, with assisting by domain-specific algorithm (e.g., Monte Carlo tree search in Go game \cite{silver2017mastering}; GrabCut in image segmentation \cite{khoreva2017simple}), training samples/labels can be generated to iteratively or recursively update the neural network parameters to achieve better performance.

In this paper, we propose a very weakly supervised approach for LAM cyst detection and segmentation. 
As shown in Fig. \ref{figannotation}, the detection and segmentation of cysts is a challenging task due to the large number of cysts, greatly variation of cyst sizes, severe touching of cysts, inconsistent image quality, and image noise and motion artifact, etc. 
Moreover, it is infeasible to obtain manual segmentation on LAM studies.

Our method, differs from weakly supervised methods, can automatically learn from medical images without any manual pixel- \cite{yang2017suggestive}, sparse- \cite{khoreva2017simple}, or image-level \cite{jia2017constrained} annotation and without pre-training a segmentation network on other labeled datasets \cite{khoreva2017simple}. Starting from classic segmentation techniques, specifically unsupervised $K$-means clustering with spatial information followed by graph cuts \cite{boykov2004experimental} refinement, the initial annotation is generated and serves as labels for a segmentation network (UNet \cite{ronneberger2015u} in this paper) learning. New networks are then recursively trained with the previous network predictions as training labels. An improved segmentation network could be trained under two hypotheses: 1) deep learning might generate better predictions than the training data labels \cite{guan2017said}, and 2) better training data labels produce better predictions \cite{silver2017mastering}. Note that the value of $K$ in $K$-means clustering is the only value provided to the framework.

\section{Methods}

Given a medical image dataset without manual annotation, our method works in a self-learning manner (Fig. \ref{figframework}), where the previously generated (first by unsupervised segmentation then by segmentation networks) pixel-level annotations serve as inputs for the next level of network learning. 

\subsection{Unsupervised Segmentation}
$K$-means clustering is an unsupervised segmentation approach. By involving pixel intensity, average and median pixel intensities of a local window into a feature space, a spatial $K$-means \cite{li2012cytoplasm} classifies the image by grouping similar pixels in the feature space into clusters. The number of clusters $K$ needs to be manually set in different applications. For the cyst segmentation in CT images, we set $K$ = 3 to obtain three clusters. $c_{1}$, $c_{2}$, and $c_{3}$ are the cluster centers, indicating cyst, lung tissue, and others, respectively.

With $c_{1}$, $c_{2}$, and $c_{3}$, we construct a three-terminal graph with the energy function consisting of a data term and a pixel continuity term as in \cite{boykov2004experimental}. The data term is assigned as the squared intensity differences between pixels and the cluster centers; The pixel continuity term is 0 when two neighboring pixels values are the same, and $\delta$ otherwise ($\delta$ = 0.003 through empirical evaluation on our data). Then, max-flow algorithm \cite{boykov2004experimental} is used to optimize the energy function, and the global optimal pixel labels are obtained.

\subsection{Segmentation Network}
After obtaining the initial annotation for all the images in the dataset by using spatial $K$-means graph cuts, UNet is used as the network architecture to learn a better segmentor because of its efficiency and accuracy for medical image segmentation \cite{ronneberger2015u}. 
UNet is constituted of four layers of contraction (pooling) and four layers of expansion (up-convolution). Skip connections from contracting path to expansive path strengthen context information in higher resolution layers.

During UNet training, the inputs are raw CT images with original resolution, and the outputs are 1-channel annotations (cross-entropy loss is utilized). The training focuses on distinguishing between cysts and lung tissues and ignoring background labels. One critical problem in training UNet for medical images is that the label/class distribution can be highly imbalanced, e.g., much more positive samples than negative or vice versa. In our experiments, we use the distribution of cysts and lung tissues in the image to balance the positive and negative classes in loss function as in \cite{xie2015holistically}. We also avoid sampling empty CT slices (no cyst in the slice) in the training.

\subsection{Recursive Learning}
The trained UNet will become its own teacher -- it is applied to segment all the CT images in training set to generate a new set of pixel-level cyst labels, which will be used as the new ground truth to train a next level UNet. The network parameters of the previous UNet are transferred to initialize the next network, and a lower learning rate is used to train the next network. 
The self-learning terminates when the similarity between successive segmentation is larger than a threshold.

\section{Experimental Methods}
In this study, we evaluated our method on a LAM dataset. A total of 183 CT volumes from patients with LAM in a natural history protocol were studied. High resolution CT scans of the chest were obtained. 
The scans contained 9-13 slices and the slice thickness ranged from 1 to 1.25 mm at 3-cm intervals. Each CT slice is with 512$\times$512 pixels.

The UNet is implemented using Caffe \cite{jia2013caffe}. We train the UNet model from scratch. Three UNet models are trained progressively in the recursive framework, named as UNet-level1, UNet-level2, and UNet-level3, respectively. The initial learning rate is 1$\times$10$^{-7}$ for UNet-level1 and decreases by a factor of 10 for every next level thanks to transfer learning from previous level. Each UNet-level is trained for 50k iterations. Mini-batch of 1 image since it provides better performance (than 5, 10, etc.) in a preliminary experiment. The proposed method is tested on a DELL TOWER 7910 workstation with 2.40 GHz Xeon E5-2620 v3 CPU, 32 GB RAM, and a Nvidia TITAN X Pascal GPU of 12 GB of memory.

Our model is trained on 166 CT volumes. The remaining 17 volumes including 5 mild, 6 moderate, and 6 severe cases are left out as unseen testing data. To evaluate the segmentation performance, a medical student manually detect and segment one slice from each of the 17 testing volumes. 
The manual segmentation was tedious that it took 4 working days. Quantification metrics include Dice coefficient and absolute difference of cyst scores (ADCS).
Cyst score is defined as the percentage of lung region occupied by cysts, which is a critical clinical factor in LAM assessment \cite{yao2014sustained}.

It's worth mentioning that differing from traditional concept of training set, our model does not learn from any manual annotation from the 166 training data (which is not available), therefore, these data can also be seen as testing data for performance evaluation. Six images (from 6 CT volumes) with large ADCS between unsupervised segmentation results and UNet results are additionally selected from the 166 dataset. Manual segmentation is then conducted on these slices for evaluation of the progressive improvement of our framework. In addition, we compare our method with the cyst segmentation method in \cite{yao2014sustained} where semi-automated thresholding followed by some postprocessing techniques were used.

\section{Results}
Table \ref{17slices} shows the performance on unseen images. 15 out of the 17 images are with good image quality while 2 are noisy. Student (i.e., UNet) learning could achieve higher segmentation accuracy than its teacher (i.e., spatial $K$-means graph cut, SK-GC), but the self-improvement seems to stop at level 3.  
The same trends could be observed in Table \ref{6slices}, where the performance on images from the learning set is shown. In these 6 CT images with large ADCS between SK-GC and UNet, compared to manual annotation, UNet learning performs substantially better than SK-GC. The lower Dice of UNet in Table \ref{17slices} compared to which in Table \ref{6slices} is mainly caused by the lower Dice values from the 5 mild cases, where both SK-GC and UNet have Dice values around 60\%. Our proposed self-learning method is also more accurate than the semi-automated method \cite{yao2014sustained}.

\begin{table}[!t] 
\centering
\caption{Performance comparison on 17 unseen CT images. SK-GC: spatial $K$-means graph cuts; ADCS: absolute difference of cyst scores. Bold indicates the best results.
}
\label{17slices}
\begin{tabular}{p{3cm}p{2.2cm}p{2.2cm}}
\hline
 & Dice (\%) & ADCS (\%)\\
\hline
Semi-automated \cite{yao2014sustained} & 62.64 & 8.34 \\
SK-GC (teacher) & 74.67 & 3.71 \\
UNet-LV.1 (student) & 75.41 & 3.65 \\
\textbf{UNet-LV.2 (student)} & \textbf{75.87} & \textbf{3.38} \\
UNet-LV.3 (student) & 74.94 & 4.56 \\
\hline
\end{tabular}
\end{table}

\begin{table}[!t] 
\centering
\caption{Performance comparison on 6 CT images with large ADCS between SK-GC and UNet from learning set. SK-GC: spatial $K$-means graph cuts; ADCS: absolute difference of cyst scores. Bold indicates the best results.
}
\label{6slices}
\begin{tabular}{p{3cm}p{2.2cm}p{2.2cm}}
\hline
 & Dice (\%) & ADCS (\%)\\
\hline
Semi-automated \cite{yao2014sustained} & 79.05 & 5.19 \\
SK-GC (teacher) & 70.39 & 11.25 \\
UNet-LV.1 (student) & 82.25 & 2.89 \\
\textbf{UNet-LV.2 (student)} & \textbf{82.65} & \textbf{1.98} \\
UNet-LV.3 (student) & 81.93 & 3.42 \\
\hline
\end{tabular}
\end{table}

   \begin{figure*}[!t]
   \begin{center}
   \begin{tabular}{c}
   \includegraphics[width=17.8cm]{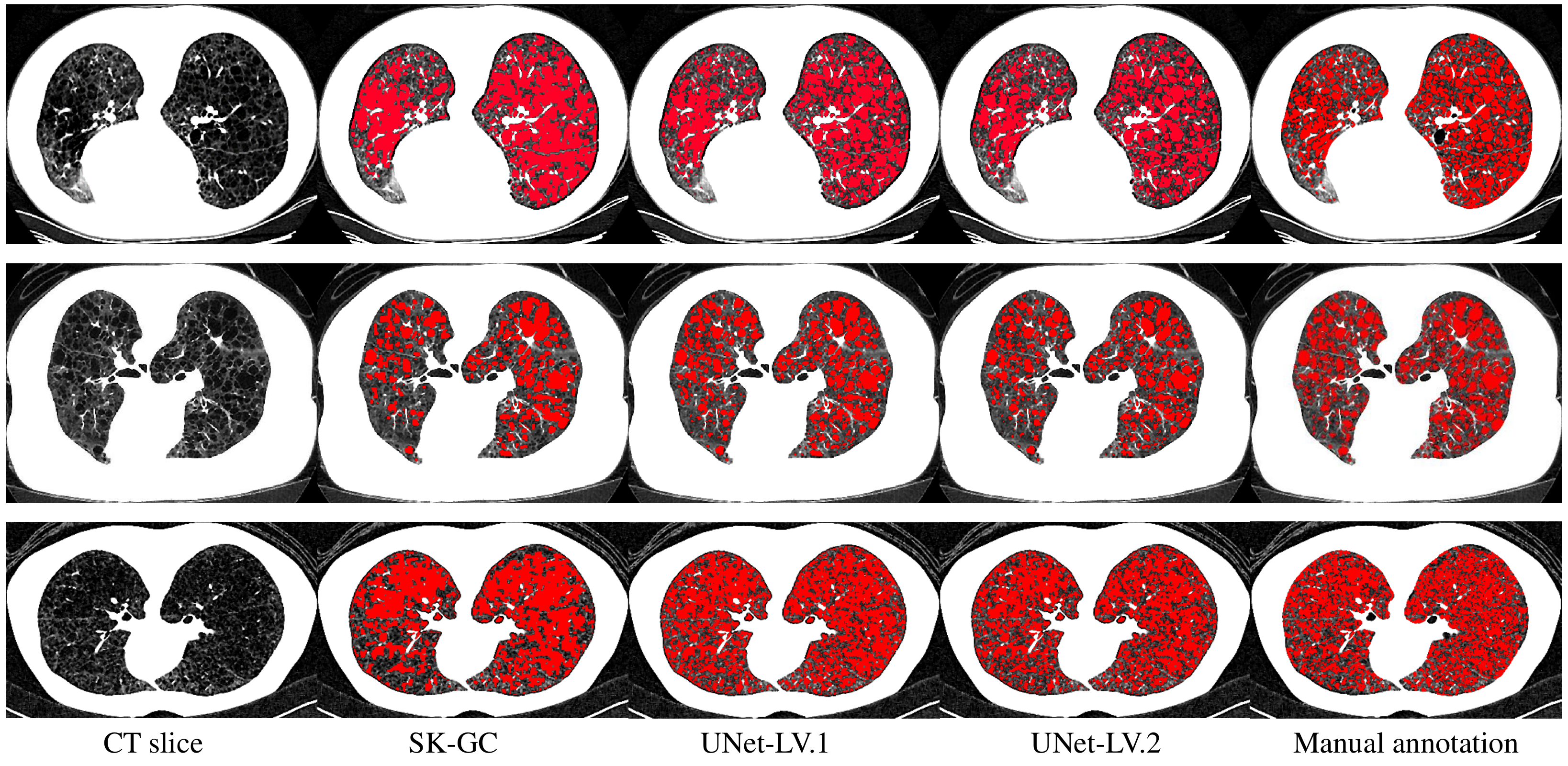}
   \end{tabular}
   \end{center}
   \caption[example]
   { \label{figresults} 
Three examples (2 good image quality and 1 noisy) show segmentation results obtained by SK-GC, UNet-level1 and UNet-level2, given manual annotation as reference. UNet-level3 is not shown due to space constraint.
}
   \end{figure*} 

Three examples in Fig. \ref{figresults} show how the proposed strategy recursively improves the segmentation performance itself. Given inaccurate segmentation provided by SK-GC, one level of UNet learning (UNet-LV.1) can already correct most oversegmentation and undersegmentation of cysts, thus achieve both higher sensitivity and higher specificity. Higher levels of UNet tend to obtain more accurate cyst boundaries especially for the overlapping cysts. The whole training process takes about 17 hours and testing is 0.13 sec./slice.

\section{Conclusions}
We report the first results of very weakly supervised learning to detect and segment cysts in lung CT images without manual annotation. By first learning from classic unsupervised segmentation, deep learning shows its potential to perform even better after a few levels of self-learning. In future work, we will extend this method to segment other medical images.

\bibliographystyle{IEEEbib}
\bibliography{strings,refs-seg}

\begin{thebibliography}{10}

\bibitem{sonka2014image}
M.~Sonka, V.~Hlavac, and R.~Boyle,
\newblock {\em Image Processing, Analysis, and Machine Vision},
\newblock Cengage Learning, 2014.

\bibitem{long2015fully}
J.~Long, E.~Shelhamer, and T.~Darrell,
\newblock ``Fully convolutional networks for semantic segmentation,''
\newblock in {\em CVPR}, 2015, pp. 3431--3440.

\bibitem{xie2015holistically}
S.~Xie and Z.~Tu,
\newblock ``Holistically-nested edge detection,''
\newblock in {\em ICCV}, 2015, pp. 1395--1403.

\bibitem{ronneberger2015u}
O.~Ronneberger, P.~Fischer, and T.~Brox,
\newblock ``U-{N}et: Convolutional networks for biomedical image
  segmentation,''
\newblock in {\em MICCAI}, 2015, pp. 234--241.

\bibitem{yao2014sustained}
J.~Yao, A.M. Taveira-DaSilva, A.M Jones, P.~Julien-Williams, M.~Stylianou, and
  J.~Moss,
\newblock ``Sustained effects of sirolimus on lung function and cystic lung
  lesions in lymphangioleiomyomatosis,''
\newblock {\em Am. J. Respir. Crit. Care Med.}, vol. 190, no. 11, pp.
  1273--1282, 2014.

\bibitem{yang2017suggestive}
L.~Yang, Y.~Zhang, J.~Chen, S.~Zhang, and D.Z. Chen,
\newblock ``Suggestive annotation: A deep active learning framework for
  biomedical image segmentation,''
\newblock in {\em MICCAI}, 2017.

\bibitem{jia2017constrained}
Z.~Jia, X.~Huang, E.~I. Chang, and Y.~Xu,
\newblock ``Constrained deep weak supervision for histopathology image
  segmentation,''
\newblock {\em IEEE TMI}, 2017.

\bibitem{guan2017said}
M.Y. Guan, V.~Gulshan, A.M. Dai, and G.E. Hinton,
\newblock ``Who said what: Modeling individual labelers improves
  classification,''
\newblock {\em arXiv preprint arXiv:1703.08774}, 2017.

\bibitem{khoreva2017simple}
A.~Khoreva, R.~Benenson, J.~Hosang, M.~Hein, and B.~Schiele,
\newblock ``Simple does it: Weakly supervised instance and semantic
  segmentation,''
\newblock in {\em CVPR}, 2017.

\bibitem{silver2017mastering}
D.~Silver, J.~Schrittwieser, K.~Simonyan, I.~Antonoglou, A.~Huang, A.~Guez,
  T.~Hubert, L.~Baker, M.~Lai, A.~Bolton, Y.~Chen, T.~Lillicrap, F.~Hui,
  L.~Sifre, G.~Van Den~Driessche, T.~Graepel, and D.~Hassabis,
\newblock ``Mastering the game of {G}o without human knowledge,''
\newblock {\em Nature}, vol. 550, pp. 354--359, 2017.

\bibitem{boykov2004experimental}
Y.~Boykov and V.~Kolmogorov,
\newblock ``An experimental comparison of min-cut/max-flow algorithms for
  energy minimization in vision,''
\newblock {\em TPAMI}, vol. 26, no. 9, pp. 1124--1137, 2004.

\bibitem{li2012cytoplasm}
K.~Li, Z.~Lu, W.~Liu, and J.~Yin,
\newblock ``Cytoplasm and nucleus segmentation in cervical smear images using
  radiating gvf snake,''
\newblock {\em Pattern Recognition}, vol. 45, no. 4, pp. 1255--1264, 2012.

\bibitem{jia2013caffe}
Y.~Jia,
\newblock ``Caffe: An open source convolutional architecture for fast feature
  embedding,''
\newblock {\em http://caffe.berkeleyvision.org/}, 2013.

\end{thebibliography}

\end{document}